\documentclass{article}

\usepackage{arxiv}

\usepackage[utf8]{inputenc} 
\usepackage[T1]{fontenc}    
\usepackage{hyperref}       
\usepackage{url}            
\usepackage{booktabs}       
\usepackage{amsmath}   
\usepackage{amssymb}   
\usepackage{amsfonts}   
\usepackage{nicefrac}       
\usepackage{cite}
\usepackage{tikz}
\usepackage{caption}
\usepackage{subcaption}
\usepackage{float}
\usepackage{times}
\usepackage{algorithmic}
\usepackage{epsfig} 
\usepackage{mathptmx} 
\usepackage{soul}

\definecolor{olivegreen}{rgb}{0,0.6,0}

\title{Using Soft Actor-Critic \\ for Low-Level UAV Control}

\author{
  Gabriel M. Barros \qquad\qquad Esther L. Colombini\\
  Laboratory of Robotics and Cognitive Systems (LaRoCS)\\  Institute of Computing, Unicamp\\ 
  Campinas, S\~ao Paulo, Brazil \\
  \texttt{gabrielmoraesbarros@gmail.com; esther@ic.unicamp.br}
   \And
  }

\begin{document}
\maketitle

\begin{abstract}
Unmanned Aerial Vehicles (UAVs), or drones, have recently been used in several civil application domains from organ delivery to remote locations to wireless network coverage. These platforms, however, are naturally unstable systems for which many different control approaches have been proposed. Generally based on classic and modern control, these algorithms require knowledge of the robot's dynamics. However, recently, model-free reinforcement learning has been successfully used for controlling drones without any prior knowledge of the robot model. In this work, we present a framework to train the Soft Actor-Critic (SAC) algorithm to low-level control of a quadrotor in a go-to-target task. All experiments were conducted under simulation. With the experiments, we show that SAC can not only learn a robust policy, but it can also cope with unseen scenarios. Videos from the simulations are available in \href{https://www.youtube.com/watch?v=9z8vGs0Ri5g}{https://www.youtube.com/watch?v=9z8vGs0Ri5g} and the code in
\url{https://github.com/larocs/SAC\_uav}
\end{abstract}

\keywords{
Deep Reinforcement Learning \and Quadrotor \and  UAV \and Control}

\section{Introduction}

While Robotics has shaped the way that the world manufactures its goods nowadays, it was mainly due to Optimum Control, with restricting tolerances and, predominantly, careful tuning of an expert. Although robotic applications in industry are suitable for high-volume production of a few different items, it cannot change rapidly for newer tasks. As the robots employed become more complex or are naturally unstable, like humanoids or drones, the harder is. If quicker modeling is wanted, it would be more beneficial to learn control policies. 

A robot model that has achieved crescent popularity is the Unmanned aerial vehicle (UAV), also known as drones. UAVs are popular research platforms with their various military and civilian usages, such as surveillance, disaster monitoring, delivery of goods, and crop monitoring \cite{choi_inverse_2017}. Although extensive research exists in controlling UAV's, traditional methods may be insufficient to cope with changing conditions, unforeseen situations, and complex-stochastic environments \cite{zhou_vision-based_2019} required for the new generation of UAVs. In fact, most quadrotor control approaches rely on a mathematical model of the UAV and its dynamics, that is non-linear and may be inaccurate due to the model's inability to capture all aspects of the vehicle's dynamic behavior. 

Model-free Reinforcement Learning (RL) \cite{Sutton:2018} is an area of research that tries to fill this gap by learning from experience. RL aims to solve the same problem as optimal control. Still, since the state transitions dynamics is not available to the agent, the consequences of its actions have to be learned by itself while interacting with the environment, by trial-and-error. However, a typical RL approach is time-consuming as it requires a significant exploration of the state-space to derive good policies. State-of-the-art methods like Soft Actor-Critic (SAC) \cite{sac1} have been successfully employed in other robotics domains to control a variety of robot models. 

In this paper, we propose to employ SAC \cite{sac1}\cite{sac_apps} to perform low-level control of a UAV in the go-to-target task. For evaluating SAC's ability to learn a successful policy, we build a reinforcement learning framework using Coppelia Simulator (former V-REP). This reliable dynamics simulator helps ease the burden of transferring results from simulation to real experiments. To assess the policy quality, we test it on conditions for which the policy was not trained and compare it with prior results presented in the literature for similar problems.

\section{Related Work}

The literature for research in UAVs with classic and modern control is vast. Recent studies have employed Reinforcement Learning for UAV control to complement classic and modern control theory in model-based approaches \cite{Omidshafiei2015ReinforcementLQ}\cite{zhang2016learning}. RL is backed by dynamics models in these scenarios, aiming at parameter tuning, or limited to attitude control. 

For high-level control, Deep Reinforcement Learning has been applied in UAVs for Navigation, Autonomous Landing \cite{ xu_monocular_2018}\cite{polvara_autonomous_2018}\cite{sampedro_image-based_2018}\cite{wang_autonomous_2019, zhou_vision-based_2019, krishnan_air_2019, airsim2017fsr} and Target Tracking \cite{li_learning_2017}.

For low-level control, however, few works have been proposed. Zhang, et. al  \cite{zhang_learning_2016} designed a low-level control using MPC with guided-policy-search. The MPC control, mapping raw sensor data to rotor velocities, is used only in the training phase (since it is computationally expensive). Then, they employed a supervised approach to learning the final policy.

In \cite{koch_reinforcement_2019}, Koch et. al  propose an open-source high-fidelity simulation environment, GymFC. They compared a PID controller to three model-free DRL approaches PPO, TRPO, and DDPG in an attitude control task (focusing only on the propellers' thrust and the agent's angular velocities). Xu et al. \cite{xu_learning_2019} used DRL (PPO) to learn how to change from fixed-wing to multicopter setup and vice-versa.

Hwangbo et al. \cite{siegwart} proposed one of the first low-level controllers with DRL. They used DRL to control a UAV, but also a PD controller to help the training phase. They employed a model-free deterministic policy gradient approach. While deterministic policy gradients have good properties and advantages, one limitation is that it requires an expensive exploration strategy. Hence, this approach leads to very sample inefficient methods that could not be trained in more complex simulators with better dynamic models. 

Lopes et al. \cite{CANOLARS} proposed a first stochastic low-level controller based on DRL (PPO) for UAV control. The authors trained a Parrot simulated robot in Coppelia Simulator \cite{coppeliaSim} for a similar task as proposed in \cite{siegwart}. However, on-policy methods are still sampling inefficient than state-of-the-art model-free off-policy methods such as Soft Actor-Critic (SAC).  

\section{Background}
\label{sac_theory}

The Soft Actor-Critic (SAC) algorithm \cite{sac1}\cite{sac-v2}\cite{sac-v3} is based upon both actor-critic methods and maximum entropy reinforcement learning. In Standard RL we want to find the policy ($\pi$) parametrization that gives the maximum cumulative reward $\sum_{t=0}^{T} \mathbb{E}_{(s_{t},a_{t}) \mathtt{\sim} \rho_{\pi}} [r(\boldsymbol{s}_{t}, \boldsymbol{a_t})]$ over an episode of length $T$ for all possible trajectories in the environment, with $\rho_\pi(s_t)$ as the state marginal of the trajectory distribution induced by policy $\pi(a_t|s_t)$.

The Q-function $Q(s,a)$ is the expected cumulative reward after taking action $a$ at state $s$. In standard RL, a unimodal policy distribution centered at the maximum Q-value is defined, and exploration happens as it extends to neighboring actions. Since exploration is biased towards one decision, the agent keeps refining its policy and completely ignores the other decisions. If the world is dynamic, a small change in it would be enough to make this agent fail. 

Maximum entropy reinforcement learning expands the maximization problem by adding an entropy bonus to be maximized through the trajectory, encouraging exploration.  It ensures the agent explores all promising states while prioritizing the more promising ones by exponentiated Q-values, where $ \pi(a_t|s_t) \propto \exp Q(\boldsymbol{s_t}, \boldsymbol{a_t})$. In this approach, which follows a Boltzmann distribution, the Q-function serves as the negative energy. Hence, as non-zero likelihood is assigned to all actions, the agent will recognize actions that lead to solving the task  \cite{bair_sac_post}.  

A maximum entropy objective that favors stochastic policies by augmenting the objective with the expected entropy of the policy over $\rho_{\pi} (s_{t})$ to encourage exploration is given by \cite{sac-v2}:

    \begin{equation}
    \label{eq_maxent}
    J (\pi) = \sum_{t=0}^{T} \mathbb{E}_{(s_{t},a_{t}) \mathtt{\sim} \rho_{\pi}} [r(s_{t}, a_{t}) + \alpha\mathbb{[H}(\pi(\cdot | s_{t}))] 
    \end{equation}

In this case, the entropy is measured by:

\begin{equation}
        \mathbb{H}(P) = \mathbb{E}_{ x \mathtt{\sim} P}[- \log P(x)]
\end{equation}

The temperature parameter $\alpha$ controls the optimal policy stochasticity. The SAC version that we use \cite{sac1} incorporates three key ingredients: an actor-critic formulation with a policy and a value function network and an off-policy approach that allows reusing previously collected data, and entropy maximization for stability and exploration \cite{sac1}. This SAC formulation results in a considerably more stable and scalable algorithm that exceeds both the efficiency and final performance of Deep Deterministic Policy Gradient (DDPG)\cite{sac-v2}.

This algorithm, in the end, aims to lift the brittleness problem of the DRL algorithms applied to continuous high-dimensional environments while maintaining the sample efficiency of off-policy algorithms. SAC has excellent convergence properties compared to its predecessors, needing fewer samples to reach good policies and finding policies with a higher reward.

\section{System Overview}

In this section, we present our proposed Reinforcement Learning framework to address the go-to-target task. We also describe the set of experiments that address the learned policy robustness and generalization capabilities.

\subsection{Problem Formulation}

\begin{figure}[htb]
     \centering
     \includegraphics[width=0.65\textwidth]{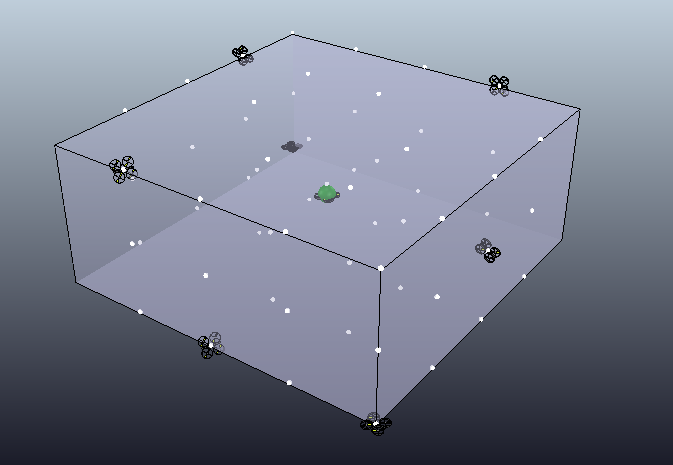}     
     \caption[Possible initial poses]{Possible initial poses for the drone that cover all quadrants of the relative agent-to-target position vector.}
     \label{fig:init}
 \end{figure}

In this work, we will train a control policy based on SAC to solve a go-to-target task. Figure \ref{fig:init} shows our simulated scene where an immobile target stands in the center (position $[x, y, z]$), with the drone dropped in the air in different positions and orientations. In this scenario, the quadrotor go-to-target task can be naturally formulated as a sequential decision-making problem under an RL framework. At each timestep, the agent observes the environment $s_t$ through its sensors (in our case, the state information given by the simulator), and acts according to policy $\pi(s)$, receiving a reward $r_t$

The agent's goal is to learn an optimal policy $\pi^{*}$ that maximizes the expected sum of discounted rewards $R_{t}$, following the SAC formulation.

\subsection{Simulation Framework}

For all experiments, our setup consists of the robotics simulator \textbf{Coppelia Simulator}\cite{coppeliaSim}, a plugin called \textbf{PyRep}\cite{pyrep}, and an environment \textbf{Environment} that access the full simulator API through Pyrep. PyRep is a C++ plugin with a Python wrapper written for the Coppelia simulator to speed up (about 20x) the agent communication with the simulator, a bottleneck for DRL training. The \textbf{Environment} corresponds to the MDP (state, action, reward) and general parameters modeled for each experiment. For all experiments, the Environment is the wrapper for the PyRep methods that interact with the simulator. The simulator then performs its integrations/calculations with a simulated version of our robot/agent, the Parrot AR Drone 2.0 \cite{siteParrot}, returning the agent's sensor readings. Figure \ref{fig:framework} shows a diagram of our framework.

We use the Bullet 2.78 physics engine and follow on Lopes et al. modified model \cite{CANOLARS} to match the Coppelia quadrotor model with the commercial quadrotor Parrot AR Drone 2.0 \cite{siteParrot}. The model was adjusted so that the main constants from AR Drone 2.0 are consistent, such as its dimensions, mass, moments of inertia, and a velocity-thrust function obtained from experiments presented in \cite{hernandez2013identification}. 

The UAV control policy outputs a value in the $[-100,100]$ range for each motor (the actions) and we apply the function $Tr(pwm) = 1.5618\cdot 10^{-4} \cdot pwm^2 + 1.0395 \cdot 10^{-2} \cdot pwm + 0.13894$ to map this value to a propeller thrust force $Tr(pwm)$  \cite{hernandez2013identification}. For our experiments, we turn off the motors internal PID.

\tikzstyle{block} = [rectangle, draw, 
    text width=6em, text centered, rounded corners, minimum height=5em]
    
\tikzstyle{line} = [draw, -latex]

\begin{figure}
\centering

\resizebox{0.8\textwidth}{!}{%

\begin{tikzpicture}[node distance = 10em, auto, thick]
    \node [block] (Agent) {Agent};
    \node [block, right of=Agent] (Environment) {Environment};
    \node [block, right of=Environment] (Pyrep) {Pyrep};
    \node [block, right of=Pyrep] (Coppelia) {Coppelia Simulator};

     \path [line] (Agent.south) --++ (0em,-2em) -| node [near start, below]{Action $a_t$} (Environment.south);
     \path [line] ([xshift=-.5cm]Environment.north) --++ (0em, 2em) -| node [near start, below]{Reward $r_t$} ([xshift=.5cm]Agent.north);
      \path [line] (Environment.north) --++ (0em, 3.5em) -| node [near start, above]{Next\_State $s_{t+1}$} ([xshift=-.5cm]Agent.north);
     
     \path [line, <->] ([xshift=.5cm]Environment.north) --++ (0,1em) -| node [near start, above] {Agent Act and Sense} (Pyrep);
     
     \path [line, <->] ([xshift=.5cm]Pyrep.north) --++ (0,1em) -| node [near start, above] {SimAPI} (Coppelia);

\draw[olivegreen, fill=olivegreen, opacity = 0.15, rounded corners] (-1.8,-2.65) rectangle(5.2,2.65);
\draw[blue, fill=blue, opacity = 0.15, rounded corners] (5.4,-2.65) rectangle(12.2,2.65);
\end{tikzpicture}
}
\caption{Reinforcement Learning framework with CoppeliaSim/Pyrep.}
\label{fig:framework}
\end{figure}
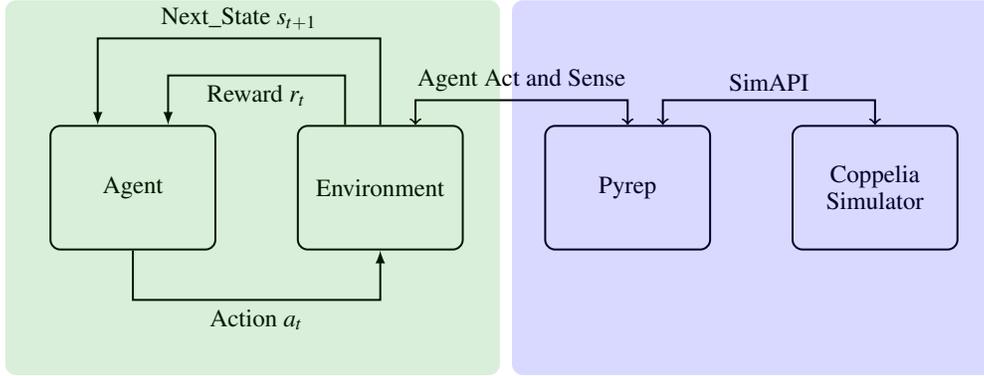


 

We carried out our experiments in a machine with the following specifications: CPU: Intel\textsuperscript{\textregistered} Core\textsuperscript{\texttrademark} i7-7700 CPU - 3.60GHz, RAM: 16GiB, GPU: NVIDIA - GeForce\textsuperscript{\texttrademark} GTX 1080 (8gb). 
 

\subsection{Reinforcement Learning Formulation}

During SAC training, in every epoch, we use the actual policy to interact with the environment and acquire trajectories to store in the replay buffer. After this, we sample $(s_t, s_{t+1}, r_t, a_t)$ from the buffer trajectories, estimating its quality ($Q$-value) for the transition $s_t  \rightarrow a_{t} \rightarrow s_{t+1}$. We use this $Q$ to weight our policy, optimizing it in the direction of the actions that increase the $Q$-value.



\subsubsection{\textbf{State}}

The state representation $S_{t}$ in our MDP (Markov Decicion Process) is composed by the relative position ($x,y,z$) and orientation ($\phi, \theta, \psi$) of the drone to the target, the relative linear ($\dot x, \dot y, \dot z$) and angular velocities ($\dot \phi, \dot \theta, \dot \psi$), the rotation matrix ($R_{11}, R_{12}, R_{13}, R_{21}, R_{22}, R_{23}, R_{31}, R_{32}, R_{33}$), and the actions taken in the previous step $a^{n}_{t-1}$ for all $n$ motors.

Although the rotation matrix has some redundant information about the agent's state, it does not contain discontinuities, and it helps prevent perceptual aliasing by removing similar representations for distinct states. The previous actions were added to represent the system dependency on the last action and to help it to infer higher-order models.


\subsubsection{\textbf{Reward-Function}}


As reward shaping is a significant element to define the quality of the policy learned by an agent, we have tested more than 20 different reward-functions, using $[x,y,z]$ positions and its derivatives (linear velocity) $[\dot x,\dot y,\dot z]$, $[\phi, \theta, \psi]$ and its derivatives (angular velocities) $[\dot \phi, \dot \theta, \dot \psi]$, death penalty (high penalty and episode termination if the agent gets too far from the target position), alive bonus (a bonus given for each time step the drone is still inside the radius of interest), the \textit{l2}-norm and standard deviation of the last action $a_{t-1}$ and integrative (accumulated) errors on $[\phi, \theta, \psi]$ and $[x,y,z]$.
    
We found that minor differences in weighting factors in the reward-function vastly outweigh differences in algorithm hyperparameters and chosen states, for example. In this work, however, we present only our best reward-function, considering the success of getting close to the target (distance reward) and robustness and stability (zeroing the angular velocity when the UAV is at the target location).

The reward function employed in this work is defined by: 

\begin{equation}
    \label{p_rew}
    r_t(s) = r_{alive} - 1.0 \|\epsilon_t(s)\| - 0.05\|\dot \phi\| - 0.05\|\dot \theta\| - 0.1\|\dot \psi\|,
\end{equation}

where $r_{alive} = 1.5$ is a constant (alive bonus) used to assure the quadrotor earns a reward for flying inside a limited region, $\epsilon_t$ is the position error (euclidean distance) between the target position and the quadrotor's position at timestep $t$.  A cost for the absolute value of the relative angular velocities ($\dot \phi, \dot \theta, \dot \psi$) is also considered. The episode terminates if the agent gets too far from the target (\textit{6.5m}).


The $r_{alive}$ term helps to improve sample efficiency and the training speed. Without it, the agent takes too long to learn. Ending the episode if the agent gets too far from the target is important especially at the beginning of the training phase when the drone practically just fall in the ground, and we would not like that these critical-failure poses appear in the training data. We applied a higher penalty to the $\dot \psi$ since it was the angular velocity component mostly responsible for the vibration (ringing effect) in our drone. 
 
\subsubsection{\textbf{Initialization}}
\label{sec:ini}

During training, the target stays always in the same pose $[x,y,z] = [0,0,1.7]$ and $[\phi, \theta, \psi] = [0,0,0]$ and we drop the drone in the air in different starting poses. The following distributions describe the possible values for the each pose component.


\begin{itemize}
    
    \item $[x, y]$ were draw from the uniform discrete distribution \\ $\mathcal{U}(-1.5, -1.0 , -0.5,  0.0 ,  0.5, 1.0, 1.5)$
    
    \item $[z]$ from $\mathcal{U}(1.2, 1.3, 1.4, 1.5, 1.6, 1.7, 1.8, 1.9, 2. , 2.1, 2.2)$
    
    \item $[\phi, \theta, \psi]$ from $\mathcal{U}(-44.69, -36.1, -26.93, -17.76,\allowbreak  -9.17,   0.0 ,  9.17,   17.76, 26.93, 36.1, 44.69)$
\end{itemize}

Figure \ref{fig:init} depicts possible initial states for the drone during training where each point represents a possible initialization for the drone with the target fixed in the center.

 \subsubsection{\textbf{Action Space}}

The action space for the problem is, for all experiments, defined by $\mathcal{A}$ = $\left\{a^{1}, a^{2}, a^{3}, a^{4}\right\}$ where each represents the value sent to each propeller. The actions range in [-100.0,100.0].

\subsection{Algorithm configuration}

For each learning process employed in this work, we use 4,000 for \textit{batch-size}, $10^6$ for \textit{buffer-size}, 0.99 for \textit{discount-factor ($\gamma$)}, $10^{-4}$ for learning rate, a (64,64) policy network with tanh as activation function and a (256,256) value and a (256,256) q-function with RELU.

We based our SAC implementation mostly on the one proposed in \cite{sac_higgisfield}, and the hyperparameters are the same as most open-source implementations of SAC. The only difference is that we choose a high value for batch-size (4000) since we want to have several starting configurations in our mini-batches in the same epoch.

    




\section{Experiments}

To evaluate the quality of the learned policy, we assess SAC's convergence and its performance in the same task we trained it: the go-to-target task with a fixed target. The target is initialized in a fixed position, located at the point $\xi_{target} = (x=0.0, y=0.0, z = 1.7)$ [m] and it stands still for the duration of the simulation training phase and evaluation. 

After that, we test our best policy's ability to generalize to other trajectories for which it was not trained (as proposed in \cite{CANOLARS}). Finally, we evaluate how robust the resulting policy is. We start the drone at each possible pose described in section \ref{sec:ini} with a fixed target. 

The time horizon of the training environment is 250 timesteps (of 0.05 seconds) for each episode, which corresponds to 12.5 seconds of flight. To test the deterministic policy and attest stability, convergence, vibration, and overall quality of flight with a moving target, we use 500 timesteps (25 seconds) in the evaluation phase.

\subsection{Immovable Target}

\begin{figure} [htb]
    \centering
        \includegraphics[width=0.45\columnwidth]{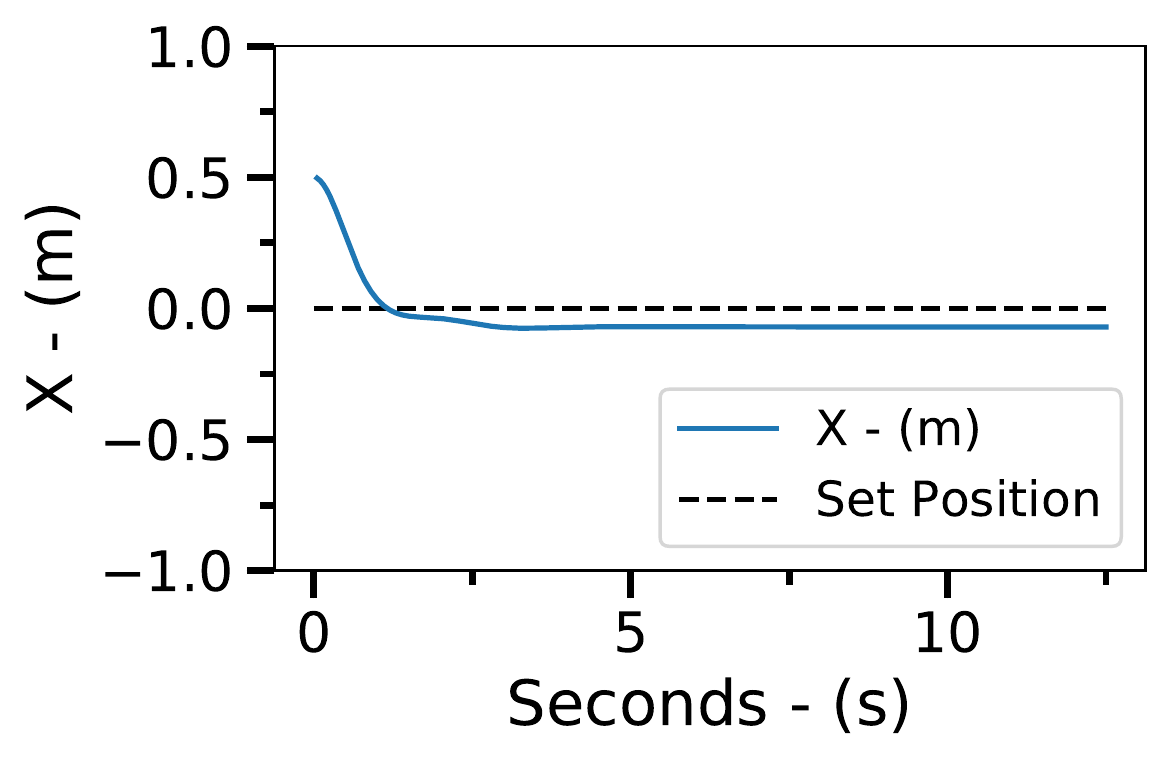}
        \hfill
        \includegraphics[width=0.45\columnwidth]{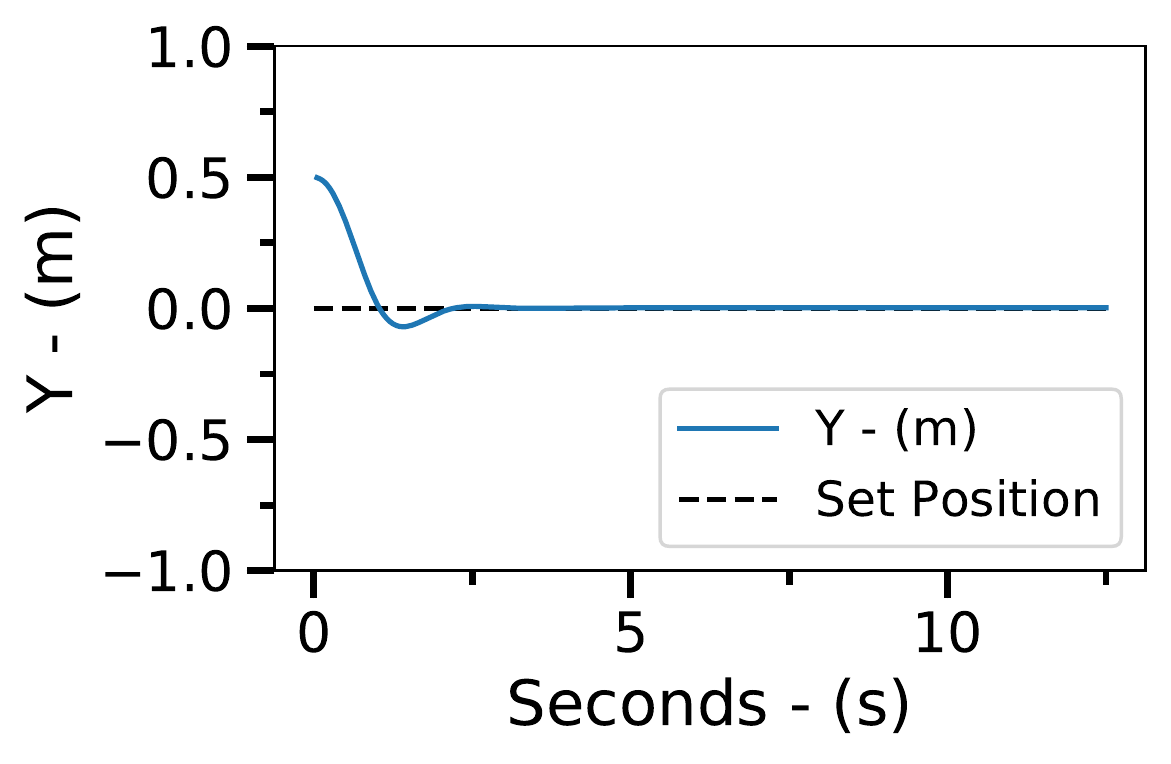}
        \includegraphics[width=0.45\columnwidth]{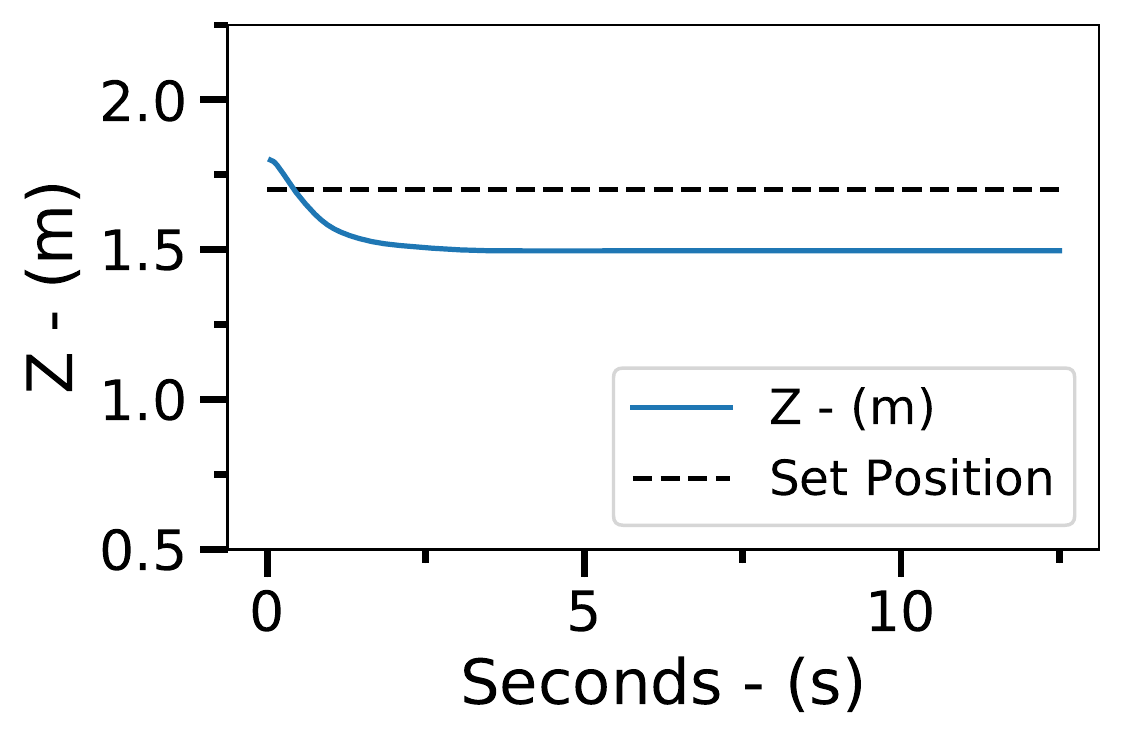}
        \hfill
        \includegraphics[width=0.45\columnwidth]{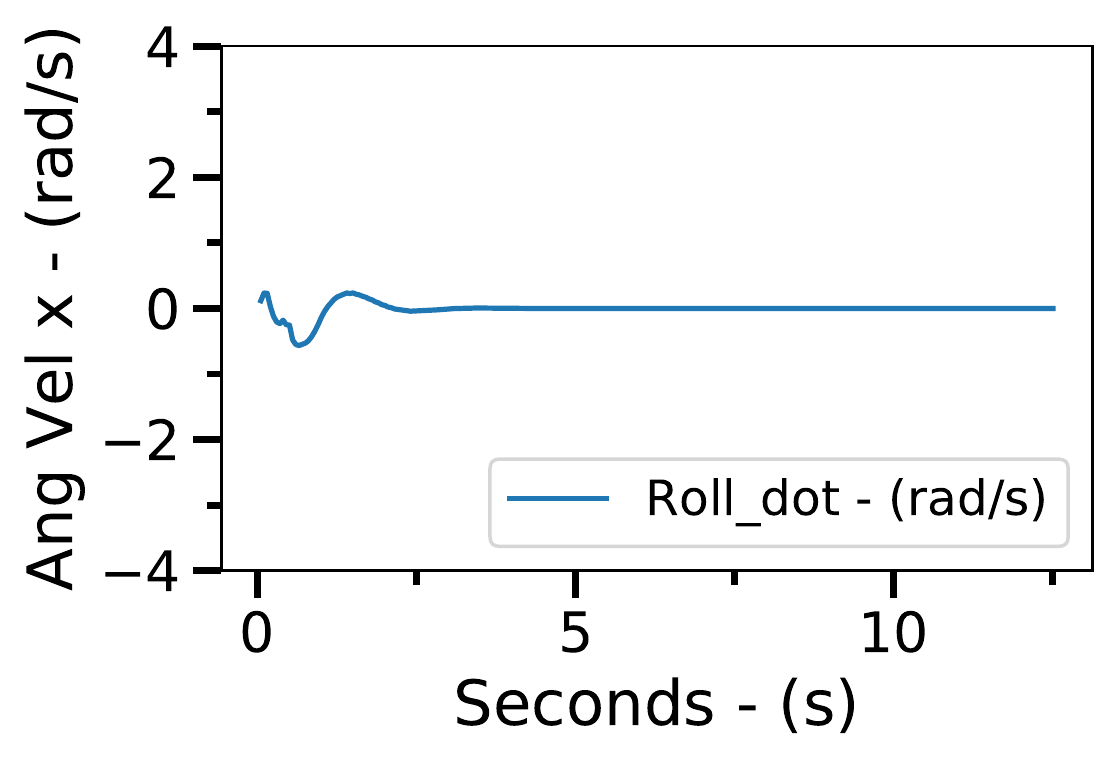}
        \includegraphics[width=0.45\columnwidth]{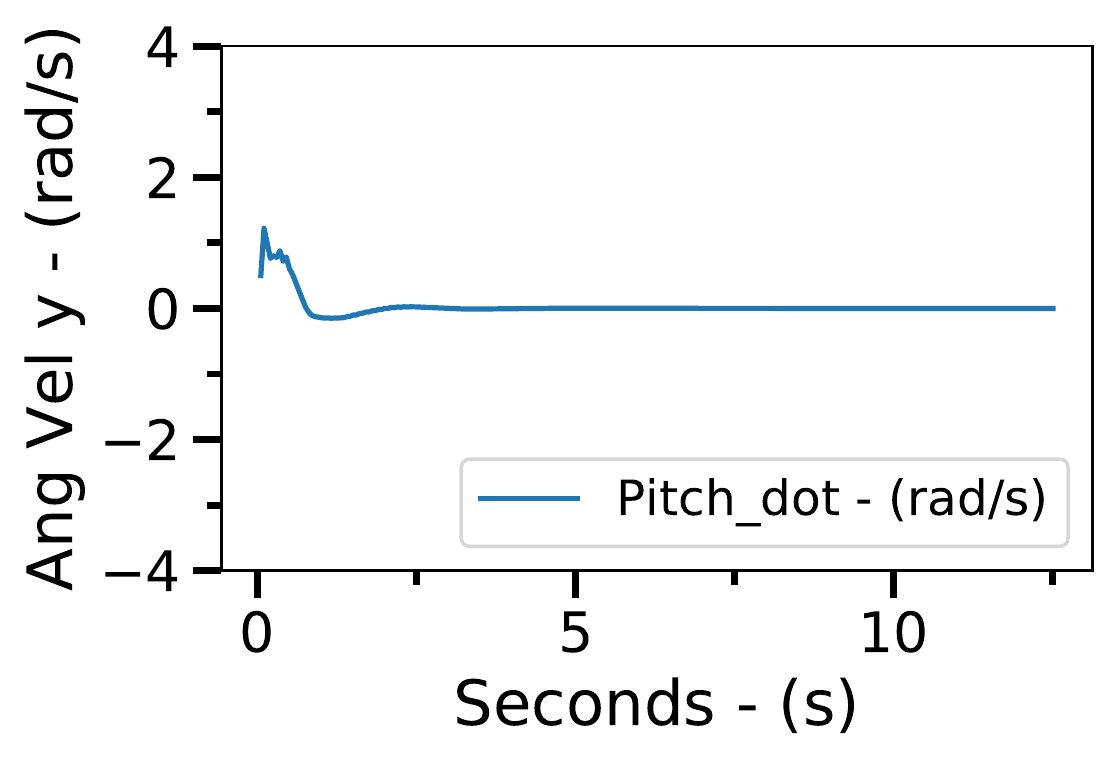}
        \hfill
        \includegraphics[width=0.45\columnwidth]{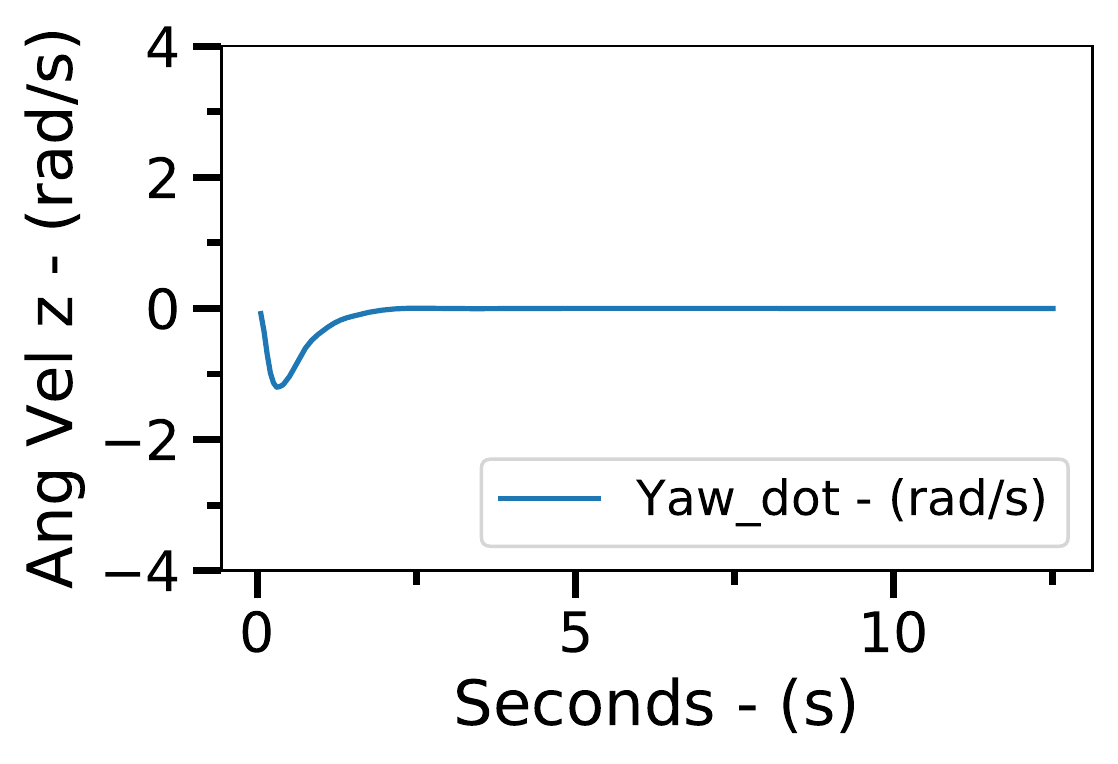}
    \caption{Drone Position ($m$) and Angular Velocities ($rad/s$).  (a) $x$ (b) $y$ (c) $z$ (d) $\dot \phi$ (e) $\dot \theta$ (f) $\dot \psi$.}%
\label{fig:sac_discretized}%
\end{figure}

\begin{figure} [htb]
    \centering
   	  \includegraphics[width=0.7\columnwidth]{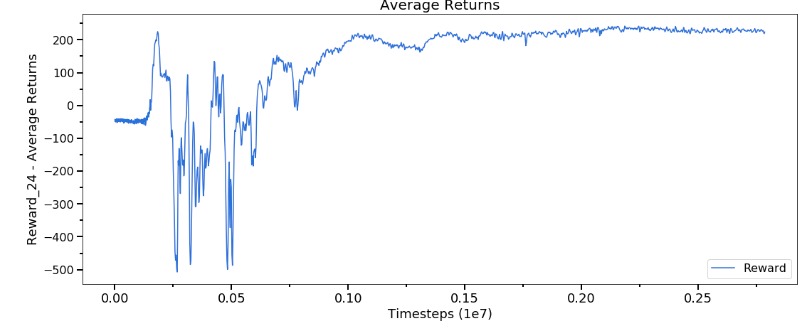} 
   	 \\
     \includegraphics[width=0.7\columnwidth]{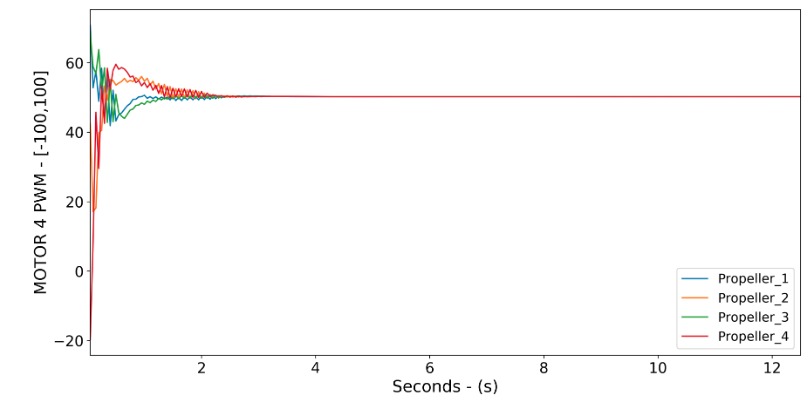}
	
    \caption{Rewards during traing and Actions from the learned policy and sent to motors for the experiment depicted in Figure \ref{fig:sac_discretized}.}%
\label{fig:pwm}%
\end{figure}

As it can be seen in Figure \ref{fig:sac_discretized}, the resulting policy achieves good stabilization capability (Figure \ref{fig:sac_discretized}) with a minor steady-error on $x$ and $z$.  The step-response with a little jump at the beginning of the episode is expected because the drone starts away from the target and is also inclined. We find that it was somewhat hard to zero these errors because the policy would alternate between getting closer to the target and achieving a stable hovering.  The PWM signals \ref{fig:pwm} sent to each of the motors showed an adequate form, as it is well-balanced among the four propellers, which is the expected attitude of a quadrotor in the situation when the drone is flying targeting a fixed point in space. In general, this is a good and stable policy with a fast-rising time with minimum overshoot and steady-state error on z-position.

\subsection{Moving Target}

The fact that the RL agent to control the quadrotor was only trained to reduce the position error to fixed targets is a known limitation. Therefore, in the remaining of this section, we focus on analyzing its attitude when subject to moving targets. For the subsequent experiments, we set the target point in movement. It now courses three different paths: a straight line, a sinusoidal and a straight path. The target moves through these paths in two different velocities: V1 = 0.2m/s and V2 = 1.5m/s

\subsubsection{Line}

\begin{figure} [htb]
\centering
      \subfloat{%
      \includegraphics[width=0.5\linewidth]{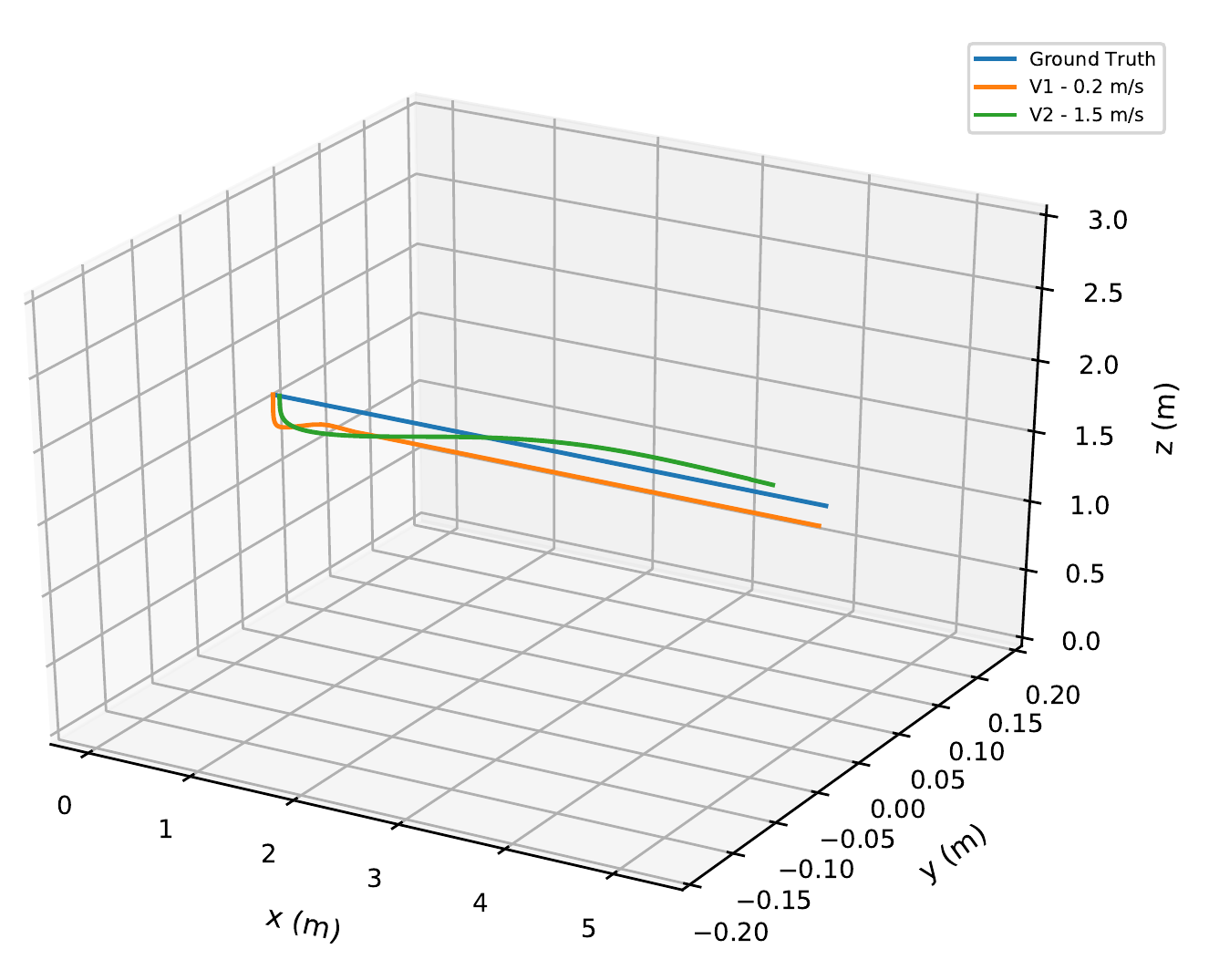}}
    \hfill
    \vspace{0.5cm}
  \subfloat{%
        \includegraphics[scale=1.40]{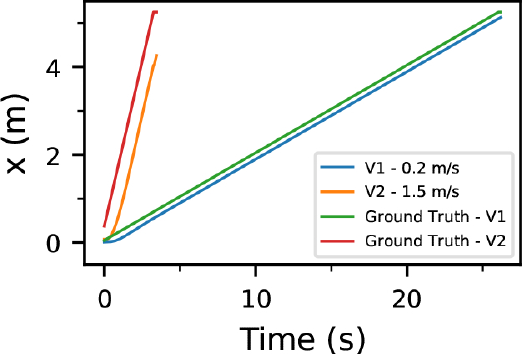}}
    \\
    
    \subfloat{%
        \includegraphics[scale=1.4]{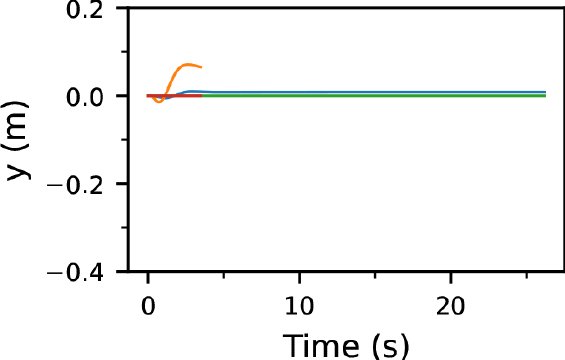}}
    \hfill
  \subfloat{%
        \includegraphics[scale=1.4]{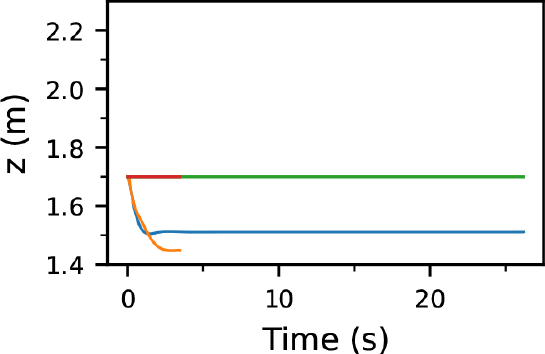}}
    \caption{SAC $\pi^{*}$ for the Line trajectory. From left to right, top to bottom. (a) SAC $\pi^{*}$ 3D position for the Line trajectory (b) $x$-Position (c) $y$-Position (d) $z$-Position}
    \label{fig:drone_line_2d_sac}%
        
\end{figure}

Figure \ref{fig:drone_line_2d_sac} depicts the drone's resulting trajectories while following the moving target at different speeds for this line trajectory. Indeed, we did not train the controller to follow the shortest path to a moving target, and it does not have any predicting capabilities. However, it can correctly follow the target by finding the shortest path to it. We can see that, as the target moves faster, the steady-state errors increase, and they shift from the positive to negative to the reference as it overshoots to reach the target.  As the drone is dropped with the beginning of the task with zero velocities in its propellers demonstrates an initial bounce  (Fig. \ref {fig:drone_line_2d_sac} c) in the trajectories caused by the step-response of the drone to the first position. Figure \ref{fig:drone_line_2d_sac} b) shows that the policy can properly follow the target that is moving in the x-position, only with the delay in the trajectory expected for such evaluations. This delay is explained because the drone is dropped without action in its propellers at the beginning of the scene while the target is already moving. Fig \ref{fig:drone_line_2d_sac} d) shows that the steady-state on z-position is almost the same in the rise and setting-time, something that shows that the policy converged to maintain the distance gap in the z-axis.

\subsubsection{Square}

\begin{figure} [htb]
    \centering
  \subfloat{
      \includegraphics[width=0.5\linewidth]{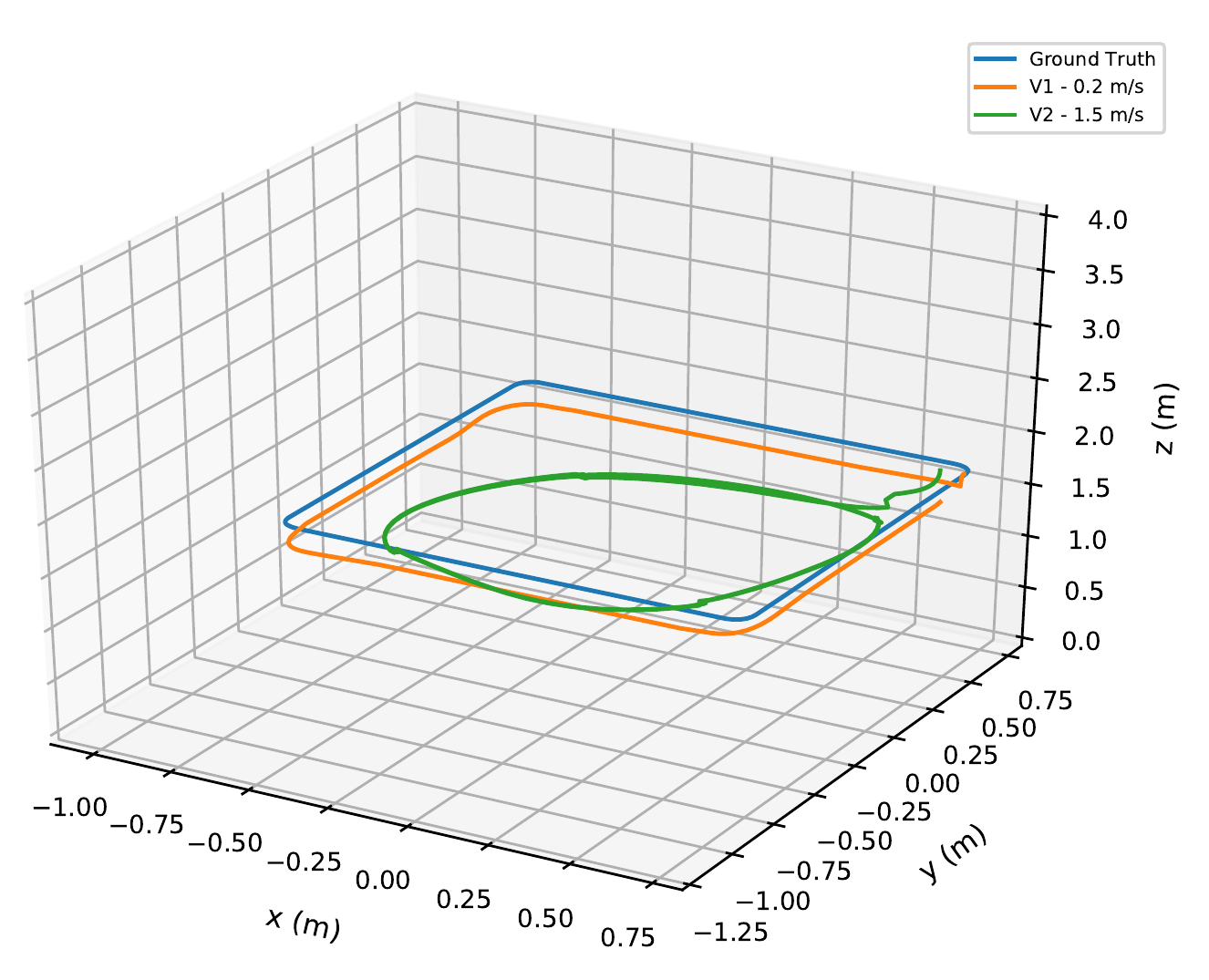}}
    \hfill
  \subfloat{%
       \includegraphics[scale=1.4]{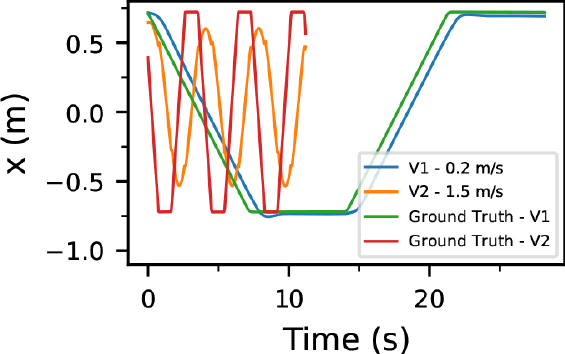}}
    \\
    \vspace{0.5cm}

    \centering
    \subfloat{%
        \includegraphics[scale=1.40]{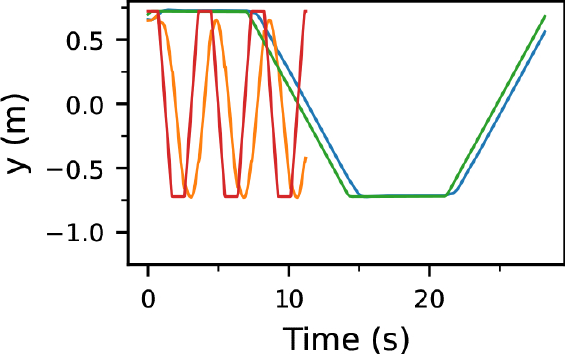}}
    \hfill
  \subfloat{%
        \includegraphics[scale=1.40]{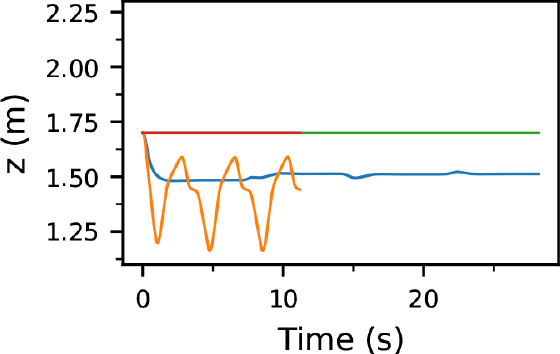}}
\caption[SAC $\pi^{*}$ for the Square trajectory]{SAC $\pi^{*}$ for the Square trajectory. From left to right, top to bottom. (a) SAC $\pi^{*}$ 3D position for the Square trajectory (b) $x$-Position (c) $y$-Position (d) $z$-Position}
\label{fig:drone_square_2d_sac}%
        
\end{figure}

As one can see in Figure \ref{fig:drone_square_2d_sac}, the square trajectory with low velocity ($V1$) is performed well by the learned policy, with almost no steady-error in $x$ and $y$. However, once again, there is a steady-state error in $z$ derived from the original policy.  Here the target moves in the x and y directions, so in b) and c), we can see how the policy behaves with the delayed trajectory but overall a clean trajectory. The agent's lag on the target can be explained because it was trained with an immobile target, so it did not learn to accelerate on an escaping target. The damps in the path ($V2$) are explained because the agent tries to go to the current target's position, leaving the square corners unvisited to minimize the current distance.

\subsubsection{Sinusoidal}

\begin{figure} [hp]

    \centering
  \subfloat{
      \includegraphics[width=0.5\linewidth]{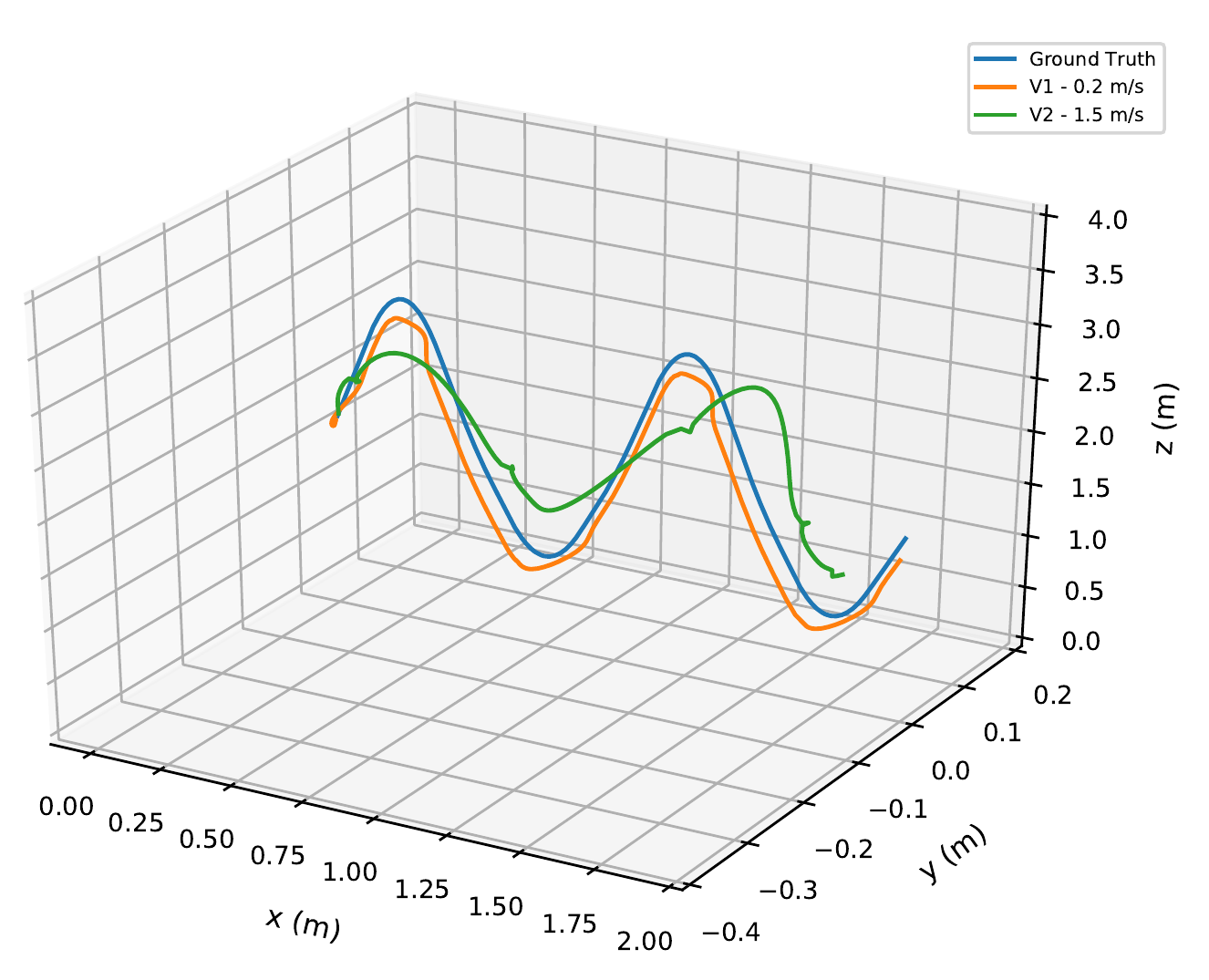}}
  \subfloat{%
        \includegraphics[scale=1.40]{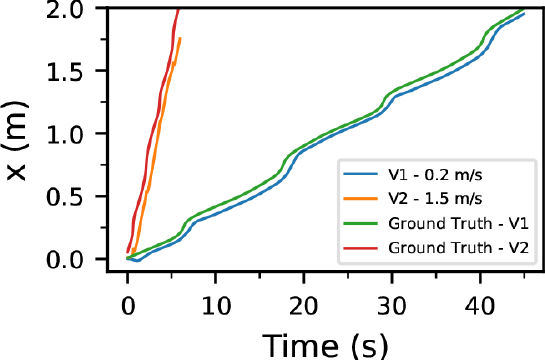}}
    \\
    \vspace{0.5cm}
    \centering
    \subfloat{%
        \includegraphics[scale=1.40]{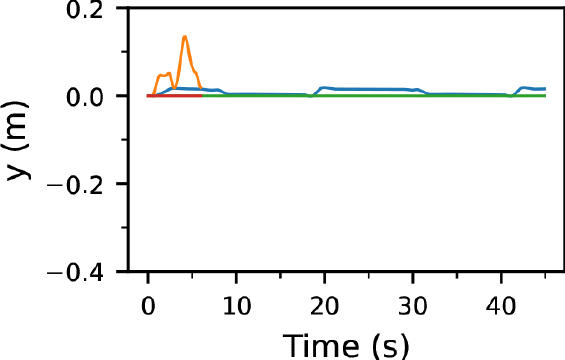}}
    \hfill
  \subfloat{%
        \includegraphics[scale=1.40]{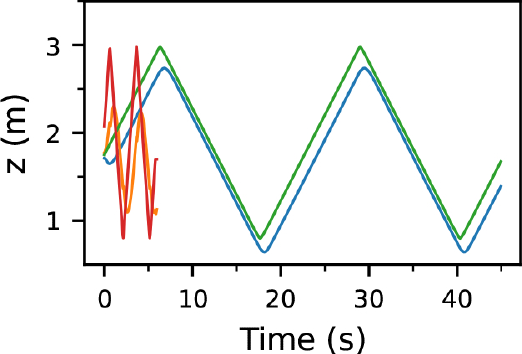}}
    \caption[SAC $\pi^{*}$ for the Sinusoidal trajectory]{SAC $\pi^{*}$ for the Sinusoidal trajectory. From left to right, top to bottom. (a) SAC $\pi^{*}$ 3D position for the Sinusoidal trajectory (b) $x$-Position (c) $y$-Position (d) $z$-Position}
    \label{fig:drone_ssenoid_2d_sac}%
\end{figure}

The SAC learned policy presented a good performance in this task (Fig. \ref{fig:drone_ssenoid_2d_sac}), especially with the slow-moving target where the trajectory is similar to the ground truth one, with differences more accentuated in the peaks and valleys of the sinusoidal curve. As expected, the curve is damped in the fast-moving trajectory, a consequence of the absence of the predicting behavior of the policy and its shortest path-driven strategy. As the target moves in $z$ and $x$, the effect of the z-position steady-state error in the trajectory is less accentuated. It can bee seen in \ref{fig:drone_ssenoid_2d_sac} c) that the error in the y-position is minimum (mainly in $V2)$, appearing basically at the beginning of the episode, vanishing later.

\subsection{Robustness tests}

To evaluate our trained policies' robustness to different testing conditions, we perform 216 episodes with extreme initialization values ($\rho_{o}$) for the drone pose while the target remains fixed. The total of possible timesteps is $216 \times 250=54000$, but the failed episodes are halt before completing 250 timesteps. The initial starting poses are drawn from the distributions described in section \ref{sec:ini}.

\begin{figure}[hp]
 \centering
\includegraphics[width=0.9\textwidth]{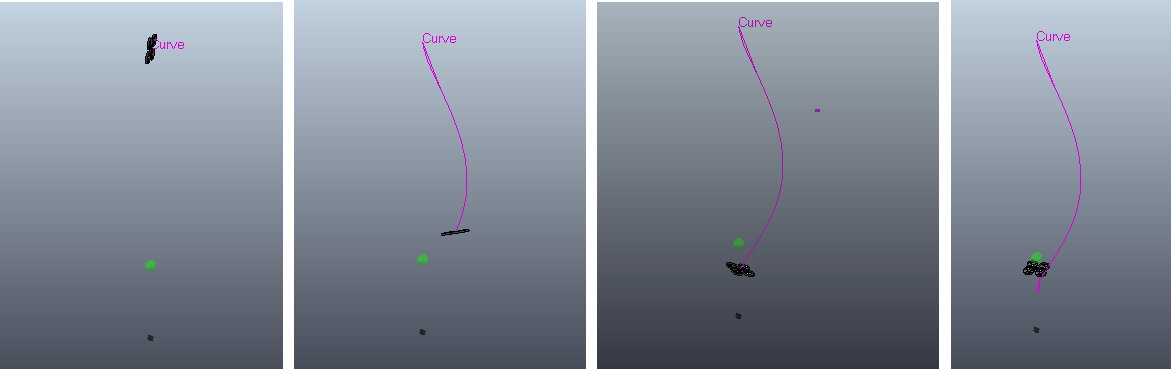}
\caption[Recovery from extreme conditions]{Recovery trajectory after being initialized with $[x,y,z] = [0,0,6]$ and $[\phi, \theta, \psi] = [0,90^{\circ},0]$ using SAC $\pi^{*s}$. The green sphere is the reference point (target).}
\label{fig:recovery}
\end{figure}

SAC performed successfully in 100\% of the runs, achieving 886.13 of median reward and 886.81 of average reward. SAC achieved a perfect score in this experiment, completing 100\% of the episodes attempted. This is a significant result considering the adverse scenarios we pushed the drone to start. Figure \ref{fig:recovery} shows an example of SAC performing over harsh untrained initial conditions. In \cite{teseGabriel}, we compare SAC results to PPO (an on-policy RL method) and to an Imitation Learning method (GAIL) learned over PPO in the same initial conditions. While PPO succeeds in 93.98\% of the tests, the GAIL approach was only able to complete 54.1\% of the runs. This fact can help us show that the maximum entropy approach on which SAC is constructed is more efficient for exploring the state-space, hence deriving more robust policies.

\section{Conclusion}


In this work,  we demonstrated that it is possible to train a reinforcement learning agent with a model-free off-policy method, specifically the Soft Actor-Critic algorithm, to obtain a policy that can perform low-level control of a simulated quadrotor. This method presents a much better sample efficiency than previously reported work for the same task, reducing the number of time steps needed for convergence from billions, such as required by the algorithm used in \cite{siegwart}, or 8 million, as presented in \cite{CANOLARS},  to around 1 million (Figure \ref{fig:sac_discretized}). This allowed us to work with a simulator with a complex dynamic engine. The method also has similar performance compared to more conservative approaches that use deterministic policy gradients, while being more robust than policy gradient methods for the same task. 

We also showed that the learned policy could successfully follow moving targets at different speeds without falls or inappropriate and dangerous behavior.  Finally, the learned policy's ability to complete 100\% of the tests in various initial conditions can indicate that SAC's state space exploration strategy can derive more robust policies, which is most desirable in naturally unstable agents like drones.
 
To advance in using DRL to controlling robotic agents in complex tasks, we must find better ways to deal with multi-objective formulations for the reward function. It is not straightforward to define a cost to some state distance metric and translate it into better-converged policy performance.  Moreover, we aim to evaluate how curriculum-learning strategies may be used to define initialization strategies that increase complexity as the drone learns.



 
\bibliographystyle{unsrt}  
\bibliography{references}

\end{document}